# TEXT CLASSIFICATION FOR AZERBAIJANI LANGUAGE USING MACHINE LEARNING AND EMBEDDINGS

Umid Suleymanov[1], Behnam Kiani Kalejahi[2,3], Elkhan Amrahov[4], Rashid Badirkhanli[5]

[1]Department of Engineering and Applied Sciences, School of Science and Engineering Khazar University, Baku, Azerbaijan, Email: umid.suleymanov@khazar.org

[2]Department of Engineering and Applied Sciences, School of Science and Engineering Khazar University, Baku, Azerbaijan, Email: bkiani@khazar.org

[3]Department of Biomedical Engineering, Faculty of Electrical and Computer Engineering, University of Tabriz, Tabriz, Iran, Email: b-kiani@tabrizu.ac.ir

[4]Department of Engineering and Applied Sciences, School of Science and Engineering Khazar University, Baku, Azerbaijan, Email: elxan.amrahov@khazar.org

[5]Department of Engineering and Applied Sciences, School of Science and Engineering Khazar University, Baku, Azerbaijan, Email: rashid.badirkhanli@khazar.org

*Abstract*— text classification systems will help to solve the text clustering problem in the Azerbaijani language. There are some text-classification applications for foreign languages, but we tried to build a newly developed system to solve this problem for the Azerbaijani language. Firstly, we tried to find out potential practice areas. The system will be useful in a lot of areas. It will be mostly used in news feed categorization. News websites can automatically categorize news into classes such as sports, business, education, science, etc. The system is also used in sentiment analysis for product reviews. For example, the company shares a photo of a new product on Facebook and the company receives a thousand comments for new products. The systems classify the comments into categories like positive or negative. The system can also be applied in recommended systems, spam filtering, etc. Various machine learning techniques such as Naive Bayes, SVM, Decision Trees have been devised to solve the text classification problem in Azerbaijani language.

*Keywords*— text classification, embedding, SVM, naive Bayes, neural network, machine learning.

## I. INTRODUCTION

### 1.1 Definition

Text classification is the task of automatically assigning one of the predefined labels to a paragraph or article. More formally, if some Di is a document of the entire set of documents D and {c1, c2, c3, …, cn} is the set of all the categories, the text classification assigns one category cj to a document di. (Suleymanov and Rustamov, 2018). In our project, each article belongs to only one category. And when a document can only belong to one category, it is called "single-label" and if the opposite is true we call this "multi-label". A "single-label" text classification task also is divided further into a "binary class" and "multi-class" classification when the document is assigned to n mutually exclusive classes. (Wang & Chiang, 2011). Text classification can help us divide up documents conceptually and has many important applications in the real world.

For example, news articles are often categorized by subject categories; academic papers are typically organized by technical domains and so on. (Yaying Qiu and et al, 2010) On the other hand, spam filtering is the widespread application of text classification.

In this kind of application of text classification, email messages are decided to be either spam or non-spam. The incoming email is automatically categorized based on its content. Language detection, analysis, and intent are based on supervised systems. Email routing and sentiment analysis are also another application of text classification. Labeled data is deployed to the machine learning algorithm and the algorithm gives the desired predefined categories. In-text classification is used labeled training data to derive a classification system and then automatically classifies unlabeled text data using the derived classifiers. Most of the data is collected from the web, especially news websites for training our data.

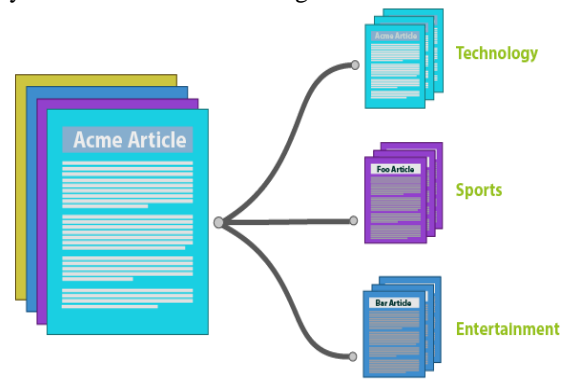

Fig. 1. Figure 1: Text Classification.

### 1.2 Purpose

As the number of digital data that is in Azerbaijani is increasing day-by-day, there is a need for classifying such data.

Especially, in the news sector, readers face such articles that they do not want to read. Assigning data to some classes can be a feasible solution to this problem. Therefore, text classification based on the topic would solve such kind of issues. Several scholarly articles and surveys have been studied during the process of research. The main target was online tutorials and surveys conducted on the area of text categorization.

As the development in technology caused an increase of resources on the web such as online documents, articles or generated text over social media in Azerbaijani language. There was a need for some way of analysing and classifying the given data for the company, organizations, and individuals. And text classification is used to solve this categorization problem for the Azerbaijani language. The developed text classification system will help to solve the text classification problems in the Azerbaijani language. There are some text-classification applications for foreign languages, but we tried to build a newly developed system to solve this problem for the Azerbaijani language. Firstly, we tried to search and find out potential practice areas that need this type of system to be applied to. The system will be useful in a lot of areas and in the future, it is expected to be used in all text related areas after the digitization process which leads to making electronic versions of handwritten documents. It will be mostly used in the news feed categorization process. News websites can automatically categorize news into classes that were defined beforehand such as sport, business, education, science, etc. The system is also used in sentiment analysis for product reviews. For example, the company shares a photo of a new product on Facebook and the company receives a thousand comments for the new product. The systems classify the comments into categories like positive or negative. In this way also help Azerbaijani companies to increase customer care, find out their weakness and development in terms of solving their mistakes. Overall, our purpose for creating text classification in Azerbaijani language is to help news websites, organizations and companies easily categorize or classify their data.

1.3 Problem Statement

For the project, to get a high percentage input, there are many machine learning algorithms that need to be applied to the project. After applying the supervised learning algorithms to the project, it needs to be compared and taken the most suitable and efficient one. Each algorithm will have its advantages. Selecting a suitable algorithm for the project does not solely determine the outcome. The text representation models and text pre-processing options also have a substantial impact on the results. When there is no prior information, BOW architecture is frequently used for text representation.

Finding the right categories which will be most appropriate for the articles is a very difficult problem. There are some conjugations between two, or event three categories which will affect the result and will decrease the preciseness of the found label. It plans to join categories too close to categories into one most suitable labels. On the other hand, stop words are another problem in the increasing percentage of the right category.

Another problem that affects the percentage of finding the right output will be the data which will be used as training for the algorithms. This data was collected from different websites and each website defines their own categories. One website categorizes news differently than another. Of course, training this kind of data will negatively affect the project output. Therefore, collected data needs to be reviewed, analysed and corrected.

## II. LITERATURE REVIEW

There is enough research on text classification. Naive Bayes is used pervasively for its speed and simple architecture ("Techniques for Improving the Performance of Naive Bayes for Text Classification", 2005). Naive Bayes classifiers utilize Bayes rule as its foundation. We can approach these problems and show that they can be solved by some simple modification. Modification can be like feature engineering, exploiting language's lexical and semantic relations using morphological resources. Some of these techniques have already applied before. ("A Comparison of Event Models for Naive Bayes Text Classification", 2017). Support Vector Machines are also widely utilized in text classification problems as Naive Bayes does and they both are supervised machine learning algorithms ("Text Categorization with Support Vector Machines with Many Relevant Features" 2006). The paper demonstrates the relative advantages of applying Support Vector Machines. First, it has high dimensional input space and few irrelevant features. For each text, the text vector consists of few entries which are not zero. Another advantage is that Support Vector Machines are a robust algorithm. Moreover, Support Vector Machines do not require any parameter because they can good parameters automatically. Thus, Support Vector Machines performs good results in text classification. Artificial neural networks have been widely applied to text categorization. ("Web Documents Categorization Using Neural Networks" 2004). Multilayer Perceptron and Decision tree algorithms are also applicable for text categorization. The paper describes the experiment of a decision tree algorithm for text categorization. The decision tree algorithm is widely used in text classification. The algorithm is a tree structure where the internal node is labeled by the term, branches represent weight and leaves represent the class. After performing experimenting Decision tree algorithms in text classification, it turns out that decision trees are capable of learning disjunctive expressions. However, it has some disadvantages such that it will not always return the globally optimal decision tree.

## III. DESIGN CONCEPTS

3.1 Description of Solutions/Approaches

As text classification is a widely encountered problem in machine learning, a lot of research has been done in this area. The text classification process is a composite process that includes pre-processing the data, training and tuning the model and at the end predicting the label of the given document from the predefined set of labels. Therefore, the accuracy of the final prediction depends not only on the model but also on problem definition and data pre-processing. For preparing the data, it is common to assign a unique number to all the words in the vocabulary and represent each document as zeros and ones where one in the given position means the document has the word in the vocabulary in exactly that position. As this representation is more easy and efficient for the computer to process, a lot of researchers use this representation for text classification. This representation is also called Bags of Words. As not all words are equally important in determining the category of the document, researchers generally use Term Frequency Times Inverse Document Frequency. Moreover, before processing the data, removing stop words and combining stem words make the calculations more efficient and accurate. Determining most of the hyper parameters and data pre-processing such as stop word removal are language-specific problems. Therefore, doing text classification for the Azerbaijani language requires a lot of novel ideas in order to achieve the desired accuracy. For example, the data set used for training the classifier is from Azeri news sites. The successful implementation of the classifier depends heavily on the data at hand. Therefore, data should be cleaned and normalized before processing which requires deep investigation of data and getting valuable insights from it. Cleaning and normalizing Azerbaijani news data is a novel problem that requires novel approaches to solve. For example, different news sites divide their articles into different categories. As a result, the news data have a lot of categories some of which are very similar to each other. Therefore, by analysing the data and categories we tried to lessen the number of categories by merging, re-assigning categories.

3.2 Naive Bayes

From the algorithmic point of view, there are several techniques to solve the current issue. The basic one is Naive Bayes which is functioning based on Bayes rule. The Naive Bayes classifier estimates the probability of new data by using the given training data. So far, as a team, we have implemented and tested this approach. The outcome appeared unsatisfactory as expected because of the working principle of Naive Bayes. Two similar words varying with a single character are perceived as two distinct strings by Naive Bayes classifier. For the next stage, we are planning to use the Support Vector Machine(SVM) for the classification of texts. The SVM integrates both dimension reduction and classification. However, it is only relevant for binary classification tasks. While using SVM, we are able to reduce the computational power and storage complexities by dividing training set into small parts and representing each as support vectors. A more advanced method is a Neural Network in which each unit will represent a single word from the training set. Neural Network produces a score rather than a probability.

Besides the algorithm, clear data has a quite high significance in order to achieve the desired accuracy. Therefore, before deciding on the algorithms, we are going to clear the current data and try to minimize the numbers of categories. Fewer numbers of categories mean the classifier is less prone to make a mistake. Moreover, even the best algorithms cannot perform well on wrongly trained data.

3.3 Word Embeddings

During image processing tasks, high-dimensional, encoded vector representations of the individual raw pixel-intensities of images are used for training machine-learning models. (Daniel Vasic, Emil Brajkovic, 2018) However, text classification techniques traditionally approach words as atomic symbols, and therefore 'mother' may be represented as id136 and 'father' as id345. These representations provide no useful information to the system regarding the interconnection of the words. Representing words as ids causes the inclusion of many zeros. Using word embedding can contribute to eliminating above mentioned problems.

3.3.1 Dataset Creation

There are 1082844 news reports in the overall data. Dataset Statistics results are for 25 February 2019. Initially, 301224 duplicate news dropped and then 781636 news remained in the overall data. Dataset statistics before the first phase cleaning is shown below.

Table 1. Sentence distribution and descriptive statistics

| Mean sentence count | 17.461187 |
|---|---|
| std | 19.988552 |
| min | 0.000000 |
| 25% percentile | 8.000000 |
| 50% percentile | 12.00000 |
| 75% percentile | 22.00000 |
| Max | 1915.000 |

Table 2. Character distribution and descriptive statistics

| Mean character count | 1466.51758 |
|---|---|
| std | 1781.99540 |
| min | 1.00000000 |
| 25% percentile | 629.000000 |
| 50% percentile | 1037.00000 |
| 75% percentile | 1693.00000 |
| Max | 186679.000 |

First Phase Cleaning involved the following: removing all news containing less than 30 characters; removing all news containing more than 10000 characters; removing all news containing less than 3 sentences; removing all news containing more than 100 sentences. 753011 news articles remained after the first phase of cleaning. The below tables summarize the dataset statistics after the first phase of cleaning. The number of all sentences (raw count) in all news articles was 12426749.

Table 3. Sentence distribution and descriptive statistics

| Mean sentence count | 16.502746 |
|---|---|
| std | 12.091614 |
| min | 4.000000 |
| 25% percentile | 8.000000 |
| 50% percentile | 13.00000 |
| 75% percentile | 22.00000 |
| Max | 99.00000 |

Table 4. Character distribution and descriptive statistics

| Mean character count | 1382.421532 |
|---|---|
| std | 1162.050153 |

| min | 59.0000000 |
|---|---|
| 25% percentile | 649.000000 |
| 50% percentile | 1050.00000 |
| 75% percentile | 1687.00000 |
| Max | 999.000000 |

The number of all characters in all news articles was 1040978620. Character distribution and descriptive statistics are illustrated below.

3.3.2 Applying Regex

A lot of JavaScript codes were observed inside the content of the news articles. These codes carry no information regarding the statistical distribution of words in sentences and therefore are meaningless for generating word embedding. Moreover, web addresses are used as reference links in some cases which also if kept can deteriorate the quality of word embedding generated at the end. Therefore, regular expressions have been implemented for getting them out of the dataset and increasing the quality of sentences. The number of news articles is 752939. Number of all sentences (dot count) in all news articles after Regex application is 9689303. Sentence distribution and descriptive statistics:

Table 5. Sentence distribution and descriptive statistics:

| Mean sentence count | 12.868643 |
|---|---|
| std | 10.661528 |
| min | 0.000000 |
| 25% percentile | 7.000000 |
| 50% percentile | 10.00000 |
| 75% percentile | 15.00000 |
| Max | 99.00000 |

Character distribution (including whitespaces) and descriptive statistics mean character count: 1299.217275 and the standard deviation is 1141.380408. The number of all characters in all news articles is 978231356. The number of words (not necessarily correct words) in all news articles is 126863549.

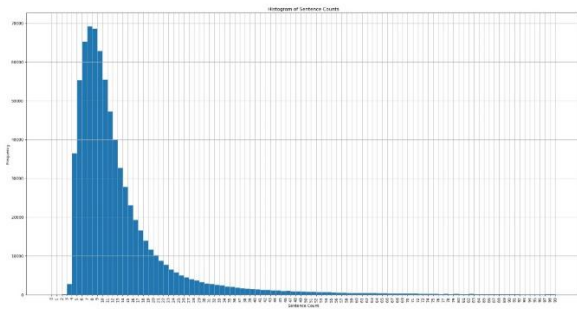

Fig 2. Word distribution in all news articles

The histogram above shows the distribution of sentences in the dataset. The vertical axis shows the number of news articles in the dataset and the horizontal axis shows the number of sentences in articles. For example, as you can see in the histogram there are 40000 news articles that contain 12 sentences. Briefly, histogram represent the most part of the articles has 5-16 sentences.

## IV. RESEARCH METHODOLOGY AND TECHNIQUES

An in-depth analysis of parameters and weights of classifiers is an essential part of the research. These parameters and weights give further insights into the classification problems. This enables us to make reasonable decisions and increase the performance results of the classifiers incrementally. The techniques used for the analysis of classifiers are as follows:

Analysing precision, f1, and other metrics of the classifier gives a lot of guidance on where the classifier suffers and how it can be fixed. Besides these metrics, there is another metrics called confusion matrix which determines how the classifier performs on the test data. More specifically it shows the number of articles our classifier classifies correctly for each category as well as the number of articles it confuses with each one of the other categories. Fig. 13 displays an instance of confusion matrix that was used for analysis purposes.

```
[[584   2   7   2   7   4]
 [  4 128   1   0   2   2]
 [  7   0 684   1  13  12]
 [  1   0   1  67   5   9]
 [  7   6  16   3 365  29]
 [  9   1  12   4  11 620]]
```

Figure 3: Confusion Matrix of SVM classifier for merged data set.

1. Architecture, Model, Diagram description

After data preprocessing, we began researching and using supervised machine learning approaches to our project so that we can optimize the prediction results. The research on the text classification was not a linear process from data preprocessing to building models and optimizing them. Rather, it was an iterative process, namely after developing the models we were analyzing the weights and coefficients to further develop our understanding of the structure of the data and to optimize the performance results of the classifier. After all these steps, the model is ready to label real-world documents on its own.

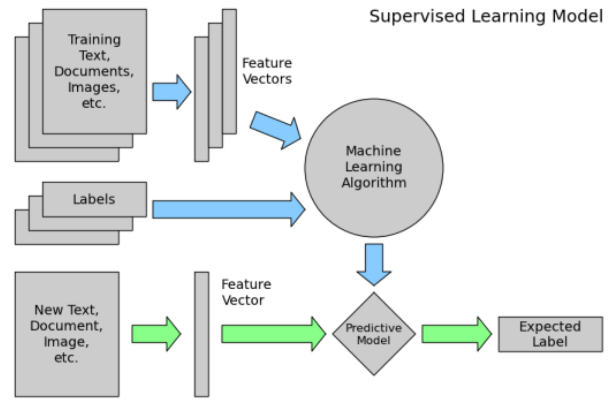

Figure 4. Architecture [7]

The project consists of two parts. The main part of the project is intended to train a model using cleaned and categorized data and use it to classify input data in the second part of the project which is web.

The image below illustrates an approach to the classification problems. Firstly, having enough data is the most essential factor in text classification problems. It is not straightforward to find clean, sanitized data for the specific problem you are solving. Therefore, you need to do some pre-processing on your data before introducing it to the classifier. You need to clear and relabel it if necessary. The data that we are using for classification has been collected from Azerbaijani news websites.

The next steps are about working with ready data. [4]

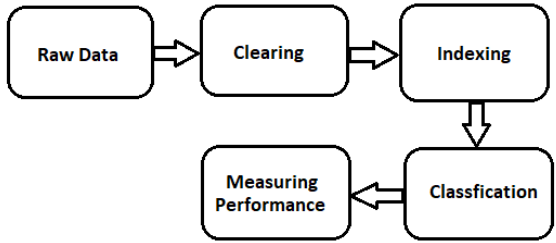

Figure 5. approach to the classification problems

2. Data Loading

Unlike other data sources, CSV and Excel files can easily be loaded and processed. The data we utilize to train the model consists of 6 columns. After the loading process, 10% of the data is kept for testing and the rest 90% is passed to the classifier.

Starting from the main part of the project, till now, different approaches have been applied, tested and evaluated.

3. Bag of Words

Bags of words are a set of various words that a document contains. The basic idea is to take any document and count the frequency of words. Based on the values of frequency, we calculate probability. It is all about how Naive Bayes works. The outcome of the code above is a bunch of tuples and integers. The way to interpret the first row of the outcome is that the word number 131607 appears only once in the first document.

```
(0, 131607)    1
(0, 116442)    1
(0, 145501)    1
(0, 194856)    1
(0, 73423)     1
(0, 28620)     1
(0, 180959)    1
(0, 194936)    1
(0, 69305)     1
(0, 24572)     1
(0, 172916)    2
```

Figure 6. bunch of tuples and integers

4. TF-IDF Vectorizer (Term Frequency-Inverse Document Frequency) Tf-idf Vectorizer is equivalent to Count Vectorizer plus Tf-idf Transformer and expresses the importance of a word in the document. By using Tf-idf Vectorizer, we can easily generate a list of the influential words for each class (category). Although Count Vectorizer is more powerful than a simple Binary Vectorizer, it has some limitations. Count Vectorizer just counts the frequency of words showing up in a document without considering the rareness or commonness of words. There is a more advanced concept, Tf-idf, which not only calculates the frequency of words, it also takes the inverse document frequency into account. The process happens in two steps: First one is about finding "tf" which is the probabilistic frequency of a word in the given document. The second one is intended to find "idf".

Table 6. Document 1(D1)

| Term | Count |
|------|-------|
| lion | 2 |
| it | 3 |
| forest | 1 |
| man | 1 |

Table 7. Document 2(D2)

| Term | Count |
|------|-------|
| cat | 1 |
| it | 3 |
| live | 1 |
| house | 1 |

$TF(\text{"IT"}, D1) = 3/7 = 0.43$

$tf(\text{"it"}, D2) = 3/6 = 0.5$
$idf(\text{"it"}, D) = \log(2/2) = 0$
$tfidf(\text{"it"}, D1) = tf(\text{"it"}, D1) \times idf(\text{"it"}, D) = 0.43 \times 0 = 0$
$tfidf(\text{"it"}, D2) = tf(\text{"it"}, D1) \times idf(\text{"it"}, D) = 0.5 \times 0 = 0$

which implies that the word "it" is not so influential in the corpus. We can go further to calculate Tf Idf of each word.

Thereby, an idf value of the word which occurs across multiple documents will below, and it will affect Tf Idf value. Low Tf Idf value of a word denotes that the word is less informative. So, Tf Idf vector does not only contain term frequencies as Count Vectorizer does, but also involves Idf values. Even though Naive Bayes classifier is powerful enough and shows satisfactory performance, it has weaknesses and it is the best approach for text classification. The first disadvantage of the Naive Bayes approach is data scarcity. At these moments, we would end up with zero while calculating probability. (We will discuss it in Testing\Verification part). Generally, there is no such rule that Naive Bayes is weaker than the Support Vector Machine(SVM). It completely depends on the size of the dataset, predefined categories, and how training data is organized.

5. Support Vector Machine

SVM is also applied as a machine learning technique in text categorization tasks. It is only suitable for binary classification tasks which mean text classification must be treated as a series of separate categorization problems. [3] At the training stage of Support Vector Machine, documents from two distinct categories are taken and SVM maps all the documents to high-dimensional space. Then, the algorithm attempts to find out a separator line which is also called hyperplane or model, between mapped points of two categories while making sure that margin is as high as possible. [5]

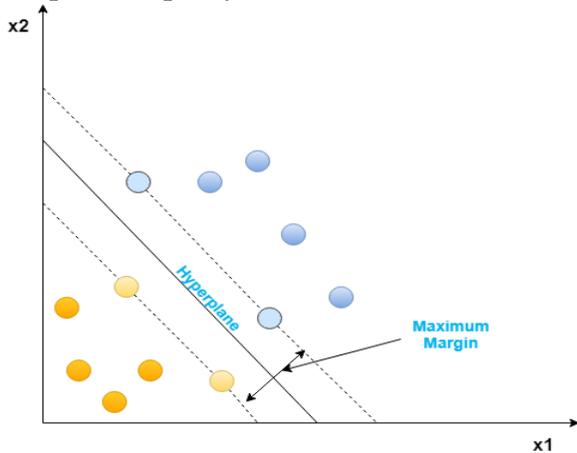

Figure 7. categorization

By implementing the Support Vector Machine, we have been able to increase the accuracy from 56.53% (Naive Bayes Classifier) to 93%. Then, to get better results than we achieved with SVM, we moved to implement another supervised machine learning algorithm, Neural Network.

6. Neural Network - Multi-Layer Perceptron

Multi-layer Perceptron is a supervised machine learning technique.

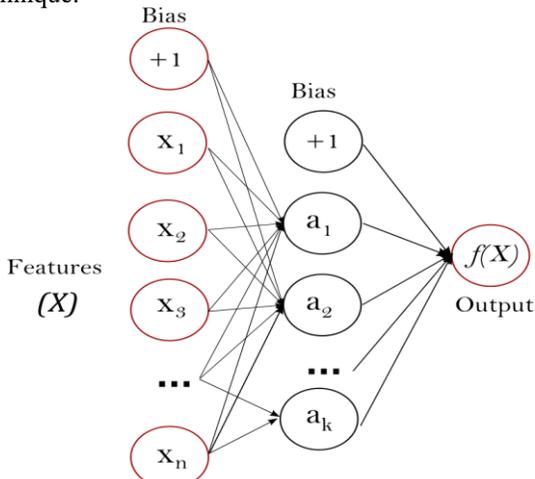

Figure 8. Multi-layer Perceptron

The code above shows how to use Multi-Layer Perceptron with "lbfgs" solver. The accuracy we achieved with Neural Network was better than Naive Bayes' outcome, however, for our dataset, SVM performs better than Neural Network implementation.

The left-most layer is called the input layer and is composed of neurons that are input features. The right-most neuron is our actual output which is classification result, category of the input document.

CONCLUSION

As predicted the Tf-idf Vectorizer performed better than the Countvectorizer because Tf-idf Vectorizer also considers the importance of a word in the document by using Tf-idf Transformer. As discussed, the Naive Bayes classifier is our initial and baseline model. The accuracy was approximately 58%, however, other research papers conclude that it can achieve more.

Additionally, determining a suitable classifier is as important as data is. After investigating the Naive Bayes approach, we shifted our attention to Support Vector Machine and we got its performance improvements. The Neural Network showed poorer performance than SVM. The scholarly articles present that the Artificial Neural Network is much more powerful than SVM, for text classification problems it cannot illustrate its full power.

Moving from classifier to web & API side of the project, we planned to run out an application on the server. First, we tried the Windows machine on Azure to setup the Flask server. However, I would say that Windows is the worst platform to run the Flask server (Apache server + WSGI module) because the integration of the WSGI module and Apache server was unsuccessful. After some effort, we moved to the Ubuntu machine, it performs well now.